\documentclass[conference]{IEEEtran}
\usepackage{times}

\usepackage[numbers]{natbib}
\usepackage{multicol}
\usepackage[bookmarks=true]{hyperref}

\usepackage{amsfonts}
\usepackage{amsmath}
\usepackage[normalem]{ulem}

\usepackage{placeins}
\usepackage{multicol}
\usepackage{afterpage}

\usepackage{amsmath,amssymb,amsfonts}
\usepackage{bm}
\usepackage{hyperref}
\usepackage{color}
\usepackage[table,xcdraw]{xcolor}
\usepackage{tabularx}
\usepackage{booktabs}
\usepackage{graphicx}
\usepackage{multirow}
\usepackage{units}
\usepackage{tikz}
\usepackage{tikz-3dplot}
\usepackage{pgfplots}
\usepackage{pgfplotstable}
\usetikzlibrary{bending, external, plotmarks, shapes, pgfplots.statistics, pgfplots.colorbrewer, positioning}
\pgfplotsset{compat=1.16} 

\newcolumntype{C}{>{\centering\arraybackslash}X}
\newcolumntype{x}[1]{>{\centering\let\newline\\\arraybackslash\hspace{0pt}}p{#1}}
\definecolor{matlab1}{rgb}{0.00000,0.44700,0.74100}
\definecolor{matlab2}{rgb}{0.85000,0.32500,0.09800}
\definecolor{matlab3}{rgb}{0.92900,0.69400,0.12500}
\definecolor{matlab4}{rgb}{0.49400,0.18400,0.55600}
\definecolor{matlab5}{rgb}{0.4660, 0.6740, 0.1880}
\definecolor{matlab6}{rgb}{0.3010, 0.7450, 0.9330}
\definecolor{matlab7}{rgb}{0.6350, 0.0780, 0.1840}
\definecolor{matlab8}{rgb}{0.8, 0.8, 0}
\definecolor{matlab9}{rgb}{0.5, 0.5, 0.5}
\definecolor{verylightgray}{rgb}{0.98,0.98,0.98}
\pgfdeclarelayer{bg}

\newcommand{\A}{\mathcal{A}}
\newcommand{\R}{\mathbb{R}}
\newcommand{\Rcal}{\mathcal{R}}

\newcommand{\Ocal}{\mathcal{O}}

\newcommand{\wfr}[0]{\ensuremath{\mathcal{W}}} %
\newcommand{\bfr}[0]{\ensuremath{\mathcal{B}}} %

\definecolor{somegray}{rgb}{0.5, 0.5, 0.5}
\newcommand{\darkgrayed}[1]{\textcolor{somegray}{#1}}
\makeatletter
\newcommand*\titleheader[1]{\gdef\@titleheader{#1}}
\AtBeginDocument{%
  \let\st@red@title\@title
  \def\@title{%
    \vskip-2.2em
    \bgroup\normalfont\large\centering\@titleheader\par\egroup
    \vskip1.0em\st@red@title}
}
\makeatother

\titleheader{\darkgrayed{This paper has been accepted for publication at Robotics: Science and Systems, 2024.}}

\title{
Demonstrating Agile Flight from Pixels \\ without State Estimation
}

\begin{document}

\author{
Ismail Geles$^*$, Leonard Bauersfeld$^*$, Angel Romero, Jiaxu Xing, Davide Scaramuzza\\ \\Robotics and Perception Group, University of Zurich, Switzerland}

\maketitle

\begin{abstract} 
Quadrotors are among the most agile flying robots. Despite recent advances in learning-based control and computer vision, autonomous drones still rely on explicit state estimation. On the other hand, human pilots only rely on a first-person-view video stream from the drone onboard camera to push the platform to its limits and fly robustly in unseen environments. To the best of our knowledge, we present the first vision-based quadrotor system that autonomously navigates through a sequence of gates at high speeds while directly mapping pixels to control commands. Like professional drone-racing pilots, our system does not use explicit state estimation and leverages the same control commands humans use (collective thrust and body rates). We demonstrate agile flight at speeds up to 40\,km/h with accelerations up to 2\,g. This is achieved by training vision-based policies with reinforcement learning (RL). The training is facilitated using an asymmetric actor-critic with access to privileged information. To overcome the computational complexity during image-based RL training, we use the inner edges of the gates as a sensor abstraction. This simple yet robust, task-relevant representation can be simulated during training without rendering images. During deployment, a Swin-transformer-based gate detector is used.
Our approach enables autonomous agile flight with standard, off-the-shelf hardware. 
Although our demonstration focuses on drone racing, we believe that our method has an impact beyond drone racing and can serve as a foundation for future research into real-world applications in structured environments.
\end{abstract}

\section*{Supplementary Material}
\noindent A narrated video with real-world experiments is available at:\\
\href{https://youtu.be/a1MSkTD-Tl8}{https://youtu.be/a1MSkTD-Tl8}

\IEEEpeerreviewmaketitle

\vspace*{-6pt}
\section{Introduction}\label{sec:introduction}

Over fifteen years after the first autonomous, vision-based quadrotor flight~\cite{bloesch2010visionbased}, 
today's most agile autonomous vision-based quadrotors still rely on explicit state estimation~\cite{LoiannoJFR2020,
shen2013aggressive,liu16icra,loianno17ral,falanga2017aggressive,mohta2018fast,lin18jfr,WangIROS21,kaufmann2020deep,Loquercio21FlightWild,kaufmann23champion}. 
This approach requires powerful, specialized hardware to perform all sensing and computation on board, as the fusion of visual and inertial information requires consistent and extremely low latencies~\cite{furgale2013calibration}. 
This starkly contrasts with professional human pilots, who instead control the drone only based on a first-person-view (FPV) video stream from the drone's onboard camera. These pilots display impressive robustness and agility as they rely on very low-level commands: collective thrust and body rates.

\begin{figure}[t!]
    \centering
    \includegraphics[width=0.95\linewidth]{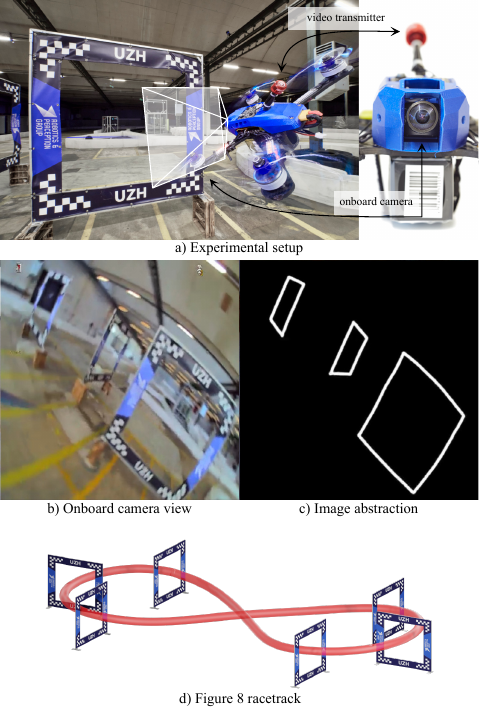}
    \vspace*{-6pt}
    \caption{Our autonomous quadrotor can fly through a racetrack purely based on images from an onboard camera, without explicit state estimation. \textbf{a)} overview of our experimental setup consisting of racing gates, a quadrotor equipped with an onboard camera, and a video transmitter, \textbf{b)} onboard view from the camera, \textbf{c)} image abstraction used by the policy, \textbf{d)} overview over the racetrack.}
    \label{fig:fig1}
    \vspace*{-18pt}
\end{figure}

The ability to control agile autonomous drones directly from image pixels without explicit state estimation and without access to IMU measurements opens tremendous possibilities. First and foremost, it enables shifting the computations to a powerful ground-station PC (or the cloud), raising the opportunity to run complex algorithms and large neural networks in real time without being constrained by the computationally-limited hardware onboard the drone. %
Secondly, not requiring specialized hardware makes a one-to-one replacement of human pilots possible, thus yielding a very scalable approach for industrial applications.

Prior work on learning agile drone flight from visual input has focused on a setting where an explicit state estimate was input to the navigation policy~\cite{kaufmann2020deep,Loquercio21FlightWild,kaufmann23champion}.
In this same setting, an autonomous drone flown by a neural-network controller trained with RL beat the human world champions of drone racing in a head-to-head race~\cite{kaufmann23champion}. 
In the context of learning a navigation policy purely from image pixels, prior works rely on the presence of a stabilizing onboard controller providing attitude and velocity estimates, severely limiting the achievable agility to below \unit[2]{m/s}~\cite{loquercio2018dronet,sadeghicad}. 

Despite the recent achievements in reinforcement learning for mobile robotics control~\cite{kaufmann23champion,lee2020learning,margolis2022rapid,miki2022anymal,fu2023learningunified,song2023reachingthelimit} and the recent advances in deep visual odometry~\cite{teed2021droid,teed2022deep,li2020deepvo,wang2017deepvo}, applying RL to learn agile-flight policies directly from pixels remains a challenging endeavor due to multiple factors.
First, RL acquires knowledge through millions of trial-and-error interactions with the environment. For vision-based RL, sample efficiency is particularly important, as (i) the dimensionality of observations will be greatly increased, making exploration more challenging, and (ii) if image rendering is required during the interactions, it will significantly raise computational costs. Therefore, the task of exploring and learning efficiently in vision-based RL becomes more challenging.
Second, in contrast with ground robots, flying robots capture images from very different viewpoints. Their ability to move freely in 3D space and the lack of constraints on the environment's appearance lead to diverse sensory information, exacerbating the difference between simulation and real-world deployment. 
Third, motion blur and rolling shutter effects degrade image quality at high speeds. 
Fourth, the simulation-to-reality gap is widened by highly nonlinear aerodynamic forces and the variability of environments, affecting both vehicle dynamics and sensory feedback. 
Fifth, quadrotors are unstable systems and the critical interdependence of perception and action necessitates rapid response times. 

\subsection{Contributions}

To our knowledge, we demonstrate the first agile vision-based autonomous quadrotor flight in the real world without explicit state estimation in a structured environment featuring known racing gate landmarks. Our drone successfully flies a racing track (Fig.~\ref{fig:fig1}d) at speeds up to \unit[40]{km/h} and accelerations up to \unit[2]{g}'s while only relying on a video stream transmitted by the drone's onboard camera (see Fig.~\ref{fig:fig1}a,b) to an offboard ground-station PC. We achieve a transmission latency of \unit[33]{ms} at a rate of \unit[60]{Hz} ($1280\times720$ resolution).

The key enablers of this breakthrough are as follows. First, directly training vision-based policies from scratch is instrumental in ensuring robust control policies that explore large regions of the observation and action spaces during training. Reinforcement learning directly from pixels is made possible by our Atari-game-inspired~\cite{mnih2015humanlevelcontrol} pixel-level abstraction (see \ref{fig:fig1}c): inner gate edges are both task-relevant and efficient to simulate during training. Second, our reinforcement learning framework leverages an asymmetric actor-critic, where the critic has access to privileged information about the full state of the drone. Third, robust gate detections in the real world are achieved by a Swin-transformer-based~\cite{liu2021swinv2} gate-detector trained on both rendered and real-world images but do not require real-world data from the deployment environment.

\section{Related Work}\label{sec:relatedwork}

\subsection{Reinforcement Learning from Pixels}
Deep reinforcement learning (RL) algorithms have achieved remarkable feats in simplified environments, such as games, providing efficient simulation platforms to benchmark the capabilities and limitations of AI agents~\cite{bellemare2013arcade}. This success is evident in surpassing human performance in Atari games~\cite{mnih2015humanlevelcontrol} and mastering complex games like Go~\cite{silver2016go}. However, these environments, primarily offering discrete actions, pose challenges for continuous control tasks. The DeepMind Control Suite~\cite{tassa2018controlsuite} addresses this gap, by presenting diverse environments for continuous control tasks. Yet, even in simulation, learning purely from pixels in these scenarios proves less sample-efficient, often yielding suboptimal performance compared to state-based learning. Recent efforts aim to improve this by refining low-dimensional visual representations alongside RL agents. Works such as SAC-AE~\cite{yarats2021sampleefficiency}, PlaNet~\cite{hafner2019planningfrompixels}, and CURL~\cite{laskin2020curl} explore techniques, including auto-encoders, future predictions, and contrastive unsupervised representations to bridge the performance-gap pixel-based to state-based policy performance gap.

Addressing the challenge of learning complex behaviors efficiently and robustly from pixels, some algorithms focus on data augmentation~\cite{laskin2020rlwithaugmenteddata,yarats2022visualcontrol}. Despite substantial progress in learning directly from pixels, these methods often cater to simulation benchmarking environments.

Diverging from the prevalent trend of simulation-centric methodologies,~\cite{wu2022daydreamer} adopts a model-based RL framework with a latent state-space model to learn from images, employing the Dreamer algorithm~\cite{hafner2020dreamer}. The learned tasks involve vision-based pick-and-place object manipulation and 2D maneuvering with a wheeled robot. The authors emphasize the challenges of mastering real-world application complexities with pixel-based control as already their quadrupedal robot relies on explicit state information. 

Sensorimotor policies, mapping camera observations to actions, are predominantly learned through extensive training data simulation, incorporating domain randomization, dynamics randomization, or additional pose information like joint angles~\cite{rusu2017sim2real,tobin2017domainrandomization,levine2016end2end}. Tasks of higher complexity in robotics often necessitate expert demonstrations, prompting a shift towards imitation learning (IL) methodologies as a prevalent choice, as opposed to the paradigm of learning from scratch using RL, showcased by recent works such as~\cite{fu2024mobilealoha}.

\subsection{Vision-based flight}
There has been a series of works that aim to fly a drone by using pixel-to-high-level commands from visual data~\cite{kaufmann2018DDR, loquercio2018dronet, shah2022gnm, shah2023vint, sridhar2023nomad}.
These approaches are not trained through reinforcement learning and are more focused on the general navigation task, and therefore are far from demonstrating agile flight.
In \cite{kaufmann2020deep}, the authors demonstrate low-level commands directly from feature representations in the image plane to perform aggressive flight maneuvers learning control policies via imitation learning.

Leaving the vision-based aspect aside, RL has been applied to low-level control of quadrotors \cite{koch2019reinforcement, lambert2019low} and has been demonstrated to be superior to classical methods.
In state-based agile flight, and more particularly, drone racing, RL methods have recently demonstrated high-speed flight~\cite{ferede2023end, eschmann2023learning}, even being able to outperform the state-of-the-art in model-based control for drone racing~\cite{song2023reachingthelimit}.
Furthermore, the versatility of RL-based controllers has allowed their adoption in autonomous, vision-based flight and several recent works have been showing ever-growing success.

One of the first uses of RL for quadrotor navigation is~\cite{sadeghi2016cad2rl}, where the authors train a vision-based RL policy using a CAD model to produce discrete 'forward', 'left', or 'right' velocity commands directly from pictures. However, this was only possible due to the drone's stabilizing onboard controller and VIO pipeline.
Very recently, an autonomous quadrotor combining vision-based state estimation and RL-based control has beaten the human world champions of drone racing~\cite{kaufmann23champion}. While this represents a milestone for robotics, the drone requires highly specialized hardware to run the state estimation and control onboard the vehicle and shows very limited robustness, crashing in \unit[40]{\%} of the races against human pilots. Additionally, to finetune the control policies of the racetrack, a motion-capture system is required for data collection, leading to reduced versatility.

In this work, we aim to overcome the limitation of specialized hardware and onboard computation by flying like professional human drone pilots\textemdash only from a video stream. This also increases the robustness of the system as much more complex algorithms can be run on an offboard ground station.

\section{Methodology}
We consider the task of vision-based agile quadrotor flight directly from pixels. The goal is to navigate through a sequence of gates (see Fig.~\ref{fig:fig1}) in the correct order as quickly as possible while only relying on a video stream from an onboard camera. In this section, we first describe the learned controller, its observation and action space, and its training procedure. Then, we detail the simulation environment for policy training, and finally, we discuss the gate detector, which is an essential part of successful real-world deployment.

\subsection{Neural Controller}
The neural controller has access to pixel observations from an onboard camera which are pre-processed in the form of a gate segmentation mask. Based on these observations, the controller has to infer control commands comprising a collective thrust and body rates (CTBR). By relying on the CTBR control modality, the drone's agility is maximized~\cite{kaufmann2022benchmark}. 

We consider the standard RL setting, consisting of an agent acting in an environment in discrete time steps within the Markov Decision Process (MDP) formulation. 
The MDP is described by the tuple $(\Ocal, \A, p, p_0, \Rcal, \gamma)$ with observation space $\Ocal$, action space $\A$, transition probabilities between states $p$, initial state distribution $p_0$ and rewards $\R$ with discount factor $\gamma$.

\subsubsection{Action Space}
At each timestep $t$, the policy needs to output an action $\bf a_t$, also referred to as the control command, which is four-dimensional. It consists of a mass-normalized collective thrust $c$ (the acceleration of the drone) and a body rate setpoint $\boldsymbol\omega_{\bfr,\text{ref}}$. These commands are then mapped to the individual motor speeds by a low-level controller onboard the drone. This control modality is also used by expert human pilots~\cite{kaufmann23champion, pfeiffer2021human} and, in contrast to high-level control commands such as linear velocities, ensures maximal agility~\cite{kaufmann2022benchmark}. Furthermore, the low-level controller on the drone is not required to estimate the state (e.g. position, attitude, velocity) of the vehicle with this control modality.

\begin{figure*}[ht]
    \centering
    \includegraphics[width=0.87\textwidth]{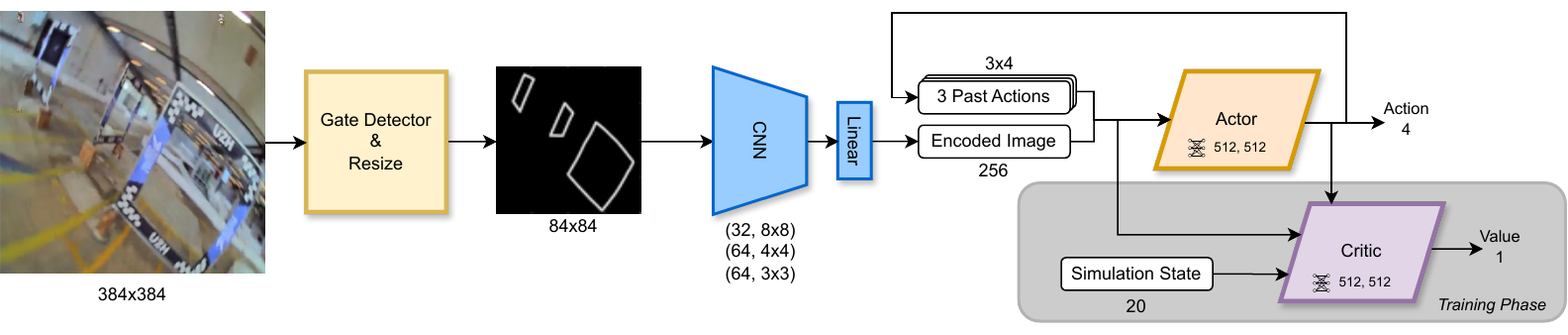}
    \caption{The architecture of our method consists of a gate detector, which is trained to segment the inner gate edges. 
    The gate detection is downsampled to a size of 84$\times$84 and given as input to the three-layer CNN acting as a shared feature extractor for the asymmetric actor-critic framework. While training, we efficiently simulate the detected gates instead of using the detector.
    Both, the actor and critic are 2 hidden layer MLPs with 512 neurons each. The actor network has access to the current image encoding and the past three actions. The critic network, which is only used when training the policy, additionally receives privileged information about the state of the simulation environment.}
    \label{fig:arch}
\end{figure*}

\subsubsection{Observation Space of the Actor}
To compute the action, the policy has access to an observation $\bm o_t$ to infer the action $\bm a_t$. In this work, the observation consists of a continuous pixel-level gate-segmentation mask (see Fig.~\ref{fig:fig1}). Here, continuous refers to a non-binary segmentation where the mask value corresponds to a confidence estimate. Our choice of observation is motivated by two aspects: first, inner gate edges are a highly task-relevant abstraction that encodes information about the task (fly through gates) as well as about the current state of the vehicle (how the gates are seen depends on where the drone is and how it is oriented). Second, during training, the inner gate edges can be obtained using perspective projection without the need to render the complete image, leading to a significant speedup. Taking inspiration from~\cite{mnih2015humanlevelcontrol}, we downsample our observations to a resolution of $84 \times 84$ pixels.

In addition to the pixel observations, the control policy has access to a history of three past actions. This enables the policy to produce smooth control commands, as the controller is not purely reactive and aware of the past commands.

\subsubsection{Observation Space of the Critic}
To train the control policies with PPO we leverage an asymmetric critic that has access to privileged information. In addition to the observations provided to the actor, the critic also has access to the full simulation state. We define this full simulation state be the 20-dimensional vector $\mathbf{s} = [\mathbf{p}, \tilde{\mathbf{R}}, \mathbf{v}, \boldsymbol{\omega}, \mathbf{i}, \mathbf{d}]$, where $\mathbf{p} \in \R^3$ is the position of the drone, $\tilde{\mathbf{R}} \in \R^6$ is a vector consisting of the first two columns of the $\mathbf{R_{\wfr\bfr}}$~\cite{zhou2019rotationmatrix}, $\mathbf{v} \in \R^3$  and $\mathbf{\omega} \in \R^3$ denote the linear and angular velocity of the drone, and $\mathbf{d} \in \R^3$ is the position of the next gate center relative to the current drone position. 

The vector $\mathbf i \in \R^2$ encodes the gate index using a sine-cosine encoding to account for the periodic nature of flying multiple laps around the track. To avoid discontinuities when a gate is passed, we use a continuous gate index. Let $i$ denote the (discrete) number of gates passed so far, and let $d = ||\mathbf d||$ denote the (scalar) distance to the next gate center. The continuous gate index $i_c$ is given by
\begin{equation}
    i_c = i + \frac{2}{1 + \exp(k\cdot d)} \;,
\end{equation}
where $k=5$ is an empirical smoothing factor. From this, the observation $\mathbf i$ can be computed as follows:
\begin{equation}
    \mathbf i = 
    \begin{bmatrix}
        \exp(-i_c) + \cos(\alpha \cdot i_c) \\
        \exp(-i_c) + \sin(\alpha \cdot i_c) \\
    \end{bmatrix} \;,
\end{equation}
where $n_G$ is the total number of gates of the racetrack, and the frequency factor is computed as $\alpha = 2\pi / n_G$. The decaying exponential summand helps distinguish between the start of a flight and the subsequent laps.

\subsubsection{Rewards}
Similar to our previous works on drone racing, we use a dense reward to guide policy learning and encode the task of navigating through the racetrack. At each timestep, the reward $r_t$ is computed as
\begin{align}
    r_t = & r_t^\text{prog} + r_t^\text{perc}  + r_t^\text{pass} - r_t^\text{cmd} - r_t^\text{crash}\; ,
\end{align}
where the individual components are calculated as follows:
\begin{alignat}{2}
    &r_t^\text{prog} &&= \lambda_1 \left( d_{t-1} - d_t \right) \nonumber\\
    &r_t^\text{perc} &&= \lambda_2 \exp \left(-\delta_\text{cam}^4\right) \nonumber\\
    &r_t^\text{cmd} &&= \lambda_3 ||\bm a_t|| + \lambda_4 ||\bm a_t - \bm a_{t-1}||^2 \\
    &r_t^\text{pass}~&&= \begin{cases}
                  1.0 - d_t, & \text{if passed gate at current timestep}\\
                  0, & \text{otherwise}
    \end{cases} \nonumber \\
    &r_t^\text{crash}~&&= \begin{cases}
                  -4.0, & \text{if $p_z<0$ or in collision with gate}\\
                  0, & \text{otherwise}
  \end{cases}\nonumber 
\end{alignat}
where $\delta_\text{cam}$ is the angle between the optical axis of the camera and the vector pointing from the drone to the next gate center. The hyperparameters {$\lambda_1=0.5$}, {$\lambda_2=0.025$}, {$\lambda_3=0.0005$}, and {$\lambda_4=0.0002$\,} are chosen empirically and trade-off speed and smoothness of a policy. The progress reward $r^\text{prog}$ encourages fast flight to maximize progress along the track, the perception reward $r^\text{perc}$ encourages orienting the camera towards the next gate to be passed, and the command smoothness penalty $r^\text{cmd}$ penalizes large control actions as well as abrupt changes in the control command. The sparse gate pass reward and collision penalty guide the policy towards safe flight.

\subsubsection{Network Architecture}
The neural controller, shown in Fig.~\ref{fig:arch}, consists of two components: A convolutional neural network (CNN) is used to extract a low-dimensional feature embedding from the high-dimensional pixel gate segmentation masks. Together with the last three actions, this embedding is used as an input to the two-layer MLP actor network. For training, the critic network has access to the CNN embedding and the full simulation state described above. 

The reasoning behind including the last three actions in the MLP is to provide the controller with some information on which commands were sent before; this short-term memory enables smoother control commands. %

\subsection{Simulation}
RL algorithms are known for being data-intensive to train.
However, data collection in real-world scenarios is often impractical, especially considering the brittle nature of quadrotors. 
To circumvent this issue, we train our control policies exclusively in simulated environments. 
This approach not only protects the hardware but also allows for a more controlled training process.

In vision-based RL, the training process involves simulating not only the dynamics of quadrotors but also generating sensory observations, such as pixel-based visual data. 
This is a key distinction from state-based RL, where such a sensor simulation is not necessary. 
When it comes to rendering images for training an RL agent, there's a trade-off to consider: opting for lower-quality rendering can speed up the training process but at the cost of less realistic images. 
On the other hand, aiming for high-fidelity, photorealistic rendering results in a reduced sim2real gap, but significantly slows down the training due to the rendering complexity.

We overcome this obstacle through the use of inner gate edges as a task-related abstraction. 
These segmentation masks can be simulated very efficiently by projecting the gate edges onto the image plane.

\subsubsection{Gate Observation}
To compute the pixel observations, the simulation environment requires knowledge about the intrinsic parameters of the camera, which are obtained using the popular Kalibr~\cite{oth2013kalibr} calibration toolbox. The lenses available for first-person-view drone racing cameras are not rectilinear but exhibit strong barrel distortion. Consequently, we use a double-sphere camera model for calibration~\cite{usenko2018doublesphere}, which is well suited to capture the distortion of wide-angle lenses. 

To simulate the pixel observations, all gate edges in view are first sorted by their distance to the camera to correctly handle occlusions. Next, each visible gate edge is discretized into 5 points which are projected into the image plane and connected with a line. By subdividing each line into multiple segments before projecting them, the resulting simulated observations correctly account for lens distortion. To robustify the policy to poor gate detections 10\% of the edge segments are corrupted and drawn at a random place in the image. 

Our simulated gate observations are very efficient to compute and take less than \unit[100]{µs} per frame, a speed unattainable with high-quality rendering.

\subsubsection{Quadrotor Dynamics} \label{sec:dynamics}
In addition to the sensor observations, the simulator must compute the dynamics of the quadrotor. This section presents a brief overview of the simulator, but the reader is referred to \cite{bauersfeld2021neurobem, kaufmann23champion} for an in-depth explanation. 
The dynamics of the quadrotor are simulated as 
\begin{align}
\label{eq:3d_quad_dynamics}
\dot{\bm{x}} =
\begin{bmatrix}
\dot{\bm{p}}_{\wfr\bfr} \\  
\dot{\bm{q}}_{\wfr\bfr} \\
\dot{\bm{v}}_{\wfr} \\
\dot{\boldsymbol\omega}_\bfr \\
\dot{\boldsymbol\Omega}
\end{bmatrix} = 
\begin{bmatrix}
\bm{v}_\wfr \\  
\bm{q}_{\wfr\bfr} \cdot \begin{bmatrix}
0 \\ \bm{\omega}_\bfr/2\end{bmatrix} \\
\frac{1}{m} \Big(\bm{q}_{\wfr\bfr} \odot (\bm{f}_\text{prop} + \bm{f}_\text{aero})\Big)+\bm{g}_\wfr  \\
\bm{J}^{-1}\big( \boldsymbol{\tau}_\text{prop} + \boldsymbol{\tau}_\text{aero}  - \boldsymbol{\omega}_\bfr \times \bm J \boldsymbol{\omega}_\bfr\big) \\
\frac{1}{k_\text{mot}} \big(\boldsymbol\Omega_\text{ss} - \boldsymbol\Omega \big)
\end{bmatrix} \; ,
\end{align} 
where $\odot$ represents quaternion rotation, $\bm{p}_{\wfr\bfr}$, $\bm{q}_{\wfr\bfr}$, $\bm{v}_{\wfr}$, and $\boldsymbol\omega_\bfr$ denote the position, orientation quaternion, inertial velocity, and bodyrates of the quadrotor, respectively. The motor time constant is $k_\text{mot}$ and the motor speeds $\Omega$ and $\Omega_\text{ss}$ are the actual and steady-state motor speeds, respectively. The matrix $\bm J$ is the quadcopter's inertia and $\bm{g}_\wfr$ denotes the gravity vector. 
The force and torque contributions of the propeller/motor unit as well as aerodynamic effects are denoted by $\bm f_\text{prop}, \boldsymbol{\tau}_\text{prop}$ and $\bm f_\text{aero}, \boldsymbol{\tau}_\text{aero}$, respectively.

To compute the force and torque contributions introduced above, we utilize either a first-principles model or a data-driven model. The first-principles model is based on blade-element-momentum (BEM) theory~\cite{bauersfeld2021neurobem, prouty1995helicopter, gill2017propeller} and while it is very accurate, it is too slow to use for training the RL controller. For this reason, we use the widespread quadratic thrust and torque propeller model with a data-driven augmentation similar to~\cite{kaufmann23champion} to obtain an accurate and computationally lightweight model. We refer to this simulation as our augmented simulator and exclusively train the controller in this setting.

\subsection{Gate Detector}
To deploy the control policies in the real world, we need a gate detector that takes as input images from the onboard camera and segments the inner gate edges. 
Compared to prior work that leverages gate detections~\cite{foehn2022alphapilot, kaufmann23champion}, this work does not rely on detected corners but uses a dense segmentation mask as an input. 
Following the advances in computer vision, the gate detector relies on a SwinTransformerV2~\cite{liu2021swinv2}. 
This architecture has proven to achieve very high performance in terms of accuracy while being very fast to compute on modern GPU architectures. 

\begin{figure}
    \centering
    \includegraphics[width=1\linewidth]{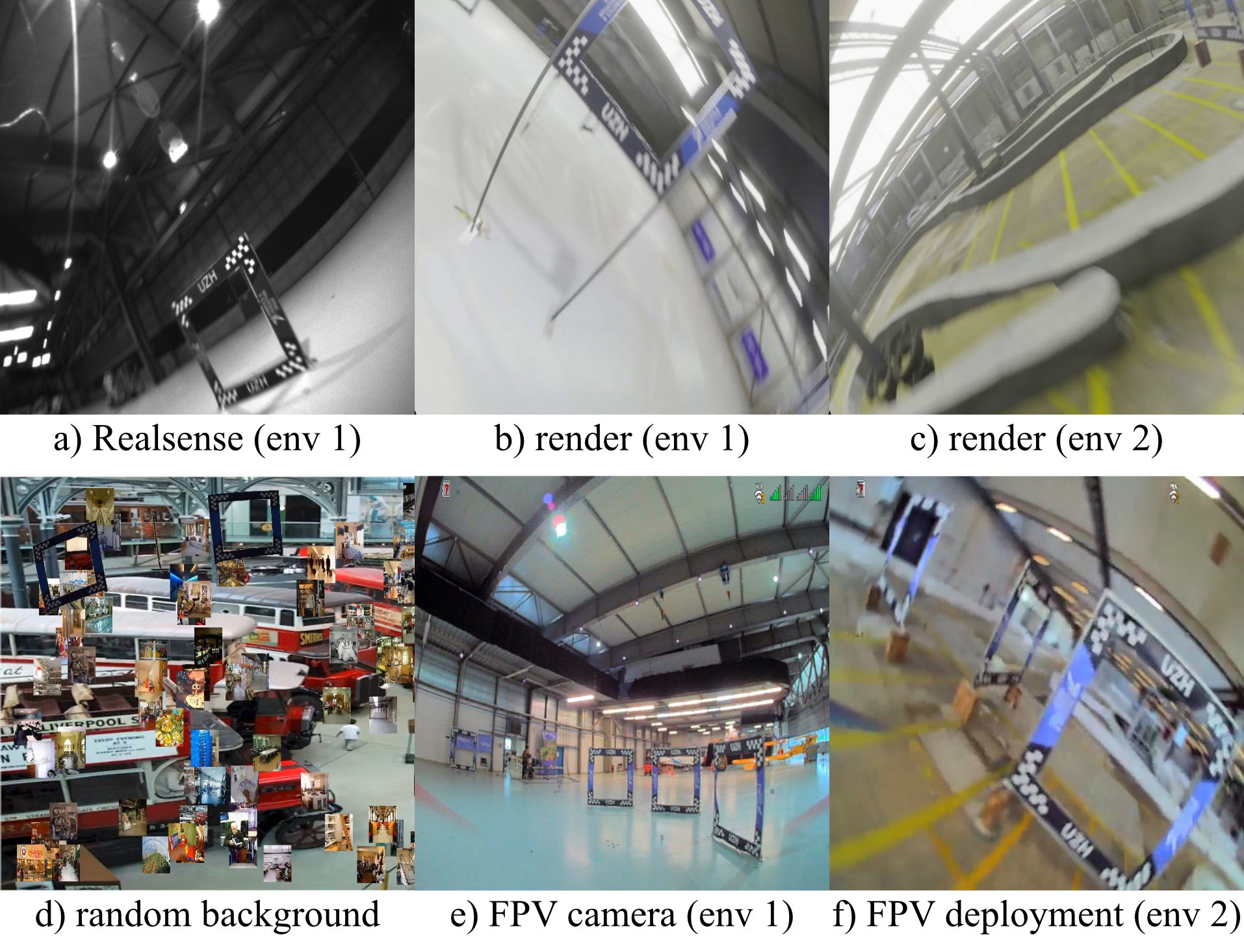}
    \caption{The gate detector is trained on data collected from real and synthetic environments; however, it has never seen real images from environment 2, in which the system is deployed (f).}
    \label{fig:gatedetector}
\end{figure}

Obtaining a robust gate detector that reliably segments gates independent of the background, adverse lighting conditions, motion blur, and rolling shutter effects is crucial for this work. To ensure this, the gate detector is trained in a supervised fashion until convergence on a diverse dataset (see Fig.~\ref{fig:gatedetector}) comprising 80,000 labeled images.  The dataset is composed as follows:
\begin{itemize}
    \item 25,000 real-world images, taken with a different camera (Intel Realsense) to the one used in this work. They are reprojected to match the FPV-camera's intrinsics (Fig.~\ref{fig:gatedetector}a)
    \item 25,000 images are photorealistic renders with strong motion blur based on digital twins of two environments (Fig.~\ref{fig:gatedetector}b,c). Environment 2 corresponds to the deployment environment.
    \item 25,000 images are generated by rendering the gates on a diverse background obtained through randomly combining images from ImageNet~\cite{russakovsky2015imagenet} together (Fig.~\ref{fig:gatedetector}d)
    \item 5,000 real-world images from the FPV-camera are included in the dataset (Fig.~\ref{fig:gatedetector}e).
\end{itemize}
Note that none of these images are obtained with the FPV camera in the deployment environment, and consequently, the gate detector is not trained on real images from the deployment environment (Fig.~\ref{fig:gatedetector}f).

For deployment, the gate detector network is implemented in C++ using TensorRT to achieve maximum performance. The inference time of the gate detector (SwinV2-B model) is only \unit[4]{ms} when deployed on an RTX 3090.

\subsection{Policy Training}
The control policy is trained through model-free reinforcement learning relying on PPO~\cite{schulman2017ppo}. The training is entirely done in simulation, where a policy is trained for 400 million environment interactions.

To obtain a policy that is capable of zero-shot transfer to the real world, it must explore a sufficiently large part of the observation space as well as exhibit robustness to a sim2real gap. To address these aspects, we employ an improved sampling strategy that relies on an initial state buffer as well as domain randomization techniques.

\subsubsection{Initial State Buffer}
Reinforcement learning can suffer from catastrophic forgetting~\cite{kaushik2021understanding}, where an agent becomes very good at performing the task and then unlearns how to recover from non-optimal states. This is caused by the policy only seeing a very narrow observation once it performs the task reliably and repeatedly. A strategy to mitigate this problem is the use of an initial state buffer~\cite{messikommer2023contrastive} from which the environment samples a state when it resets itself after an episode is terminated. 

In this work, we use a simplified version. The state buffer is used to store quadrotor states during training such that episodes are started with physically plausible, yet diverse states. We maintain a buffer of 10 possible initial states per gate. At the start of the training, the states are all initialized, such that the drone is positioned in the gate center and flies forward at a velocity of \unit[2]{m/s}. 
When the agent passes a gate, its current state is added to the state buffer if it passes the gate with a sufficient margin to the gate edges. When the environment is reset, it samples a state from the buffer and perturbs the position, attitude, velocity, and body rate randomly. This effectively prevents mode collapse and catastrophic forgetting.

\subsubsection{Domain Randomization}
Similar to the RL agents presented in previous works~\cite{song2023reachingthelimit, kaufmann23champion}, we employ domain randomization to robustify a control policy against a sim2real gap in the system dynamics. Despite having access to very accurate models, drone-to-drone variation and complex non-linear aerodynamic effects necessitate robust policies. Furthermore, domain randomization prevents a policy essentially from memorizing a sequence of control commands and forces it to react to the environment.

During training, we vary the drone dynamics by \unit[$\pm 20$]{\%} in the thrust, body drag, and inertia parameters. 
The mass is randomized by \unit[$\pm 5$]{\%}. 
The starting pose of the drone undergoes uniform sampling around the starting point, with \unit[$\pm 0.8$]{m} variations in the $x-y$ plane, \unit[$\pm 0.6$]{m} in the $z$ axis, and \unit[$\pm 20$]{deg} in attitude. The initial velocity is sampled within \unit[$\pm 0.8$]{m/s}, and initial body rates are randomized in \unit[$\pm 45$]{deg/s}. To better generalize to uncertainties in gate positions, we also randomize the gate position by \unit[$\pm 5$]{cm} in each axis ($x, y, z$). 

Therefore, the domain randomization procedure affects not only the drone's physical dynamics but also the environment conditions. It is thus fully incorporated into the full simulation state, which is used by the privileged critic and state-based policies.

\section{Experiments}\label{sec:results}

\begin{figure*}[ht]
    \centering
    \includegraphics{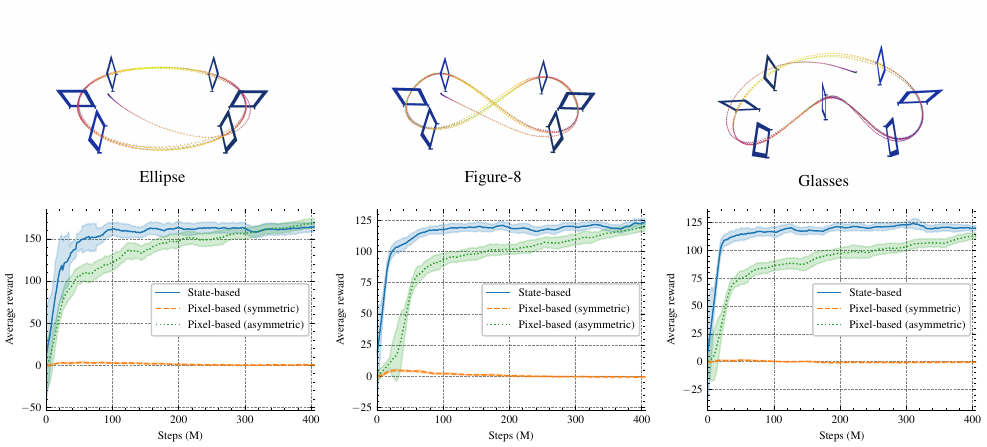}
\caption{The top row shows the three different racetracks with trajectories flown by the pixel-based policy in the augmented simulator, while the bottom row shows the reward progress during training. With the asymmetric actor-critic architecture, the average reward of our pixel-based agent converges to a similar value as the state-based agent. Excluding the privileged information for the critic network results in unsuccessful learning. The training process requires roughly 3 hours for the state-based policy, while the pixel-based policies take approximately one day to train for 400 million steps.}
\label{fig:tracks}
\end{figure*}

\begin{table*}[ht]
\centering
\caption{\textnormal{Simulation and HIL results. We compare the success rate ~(SR), mean-gate-passing-error~(MGE), and the lap-time~(LT) of our asymmetric pixel-based policy against four baselines. Three different racetracks are evaluated in simulation and HIL experiments, each with three laps. Pixel-based policies trained with our asymmetric actor-critic architecture perform significantly better than the symmetric architecture. We compare against the state-based approach of Song et al.~\cite{Song23Reaching} and a modified version, denoted by +Perc, which uses the same observations as~\cite{Song23Reaching} but with our perception-aware reward design. The best state-based result is underlined and the best pixel-based result is bold.}}
\scriptsize
\setlength{\tabcolsep}{11pt}
\begin{tabularx}{\textwidth}{ll c *9{C}}
\toprule
   &   & \multicolumn{3}{c}{BEM} & \multicolumn{3}{c}{Augmented} & \multicolumn{3}{c}{HIL} \\
        \cmidrule(lr){3-5} \cmidrule(l){6-8} \cmidrule(l){9-11} 
        & & \multicolumn{1}{c}{SR} & \multicolumn{1}{c}{MGE} & \multicolumn{1}{c}{LT}  
        & \multicolumn{1}{c}{SR} & \multicolumn{1}{c}{MGE} & \multicolumn{1}{c}{LT}  & \multicolumn{1}{c}{SR} & \multicolumn{1}{c}{MGE} & \multicolumn{1}{c}{LT} \\
 Racetrack   & Observation & \multicolumn{1}{c}{[\%]} & \multicolumn{1}{c}{[m]} & \multicolumn{1}{c}{[sec]}  
       & \multicolumn{1}{c}{[\%]} & \multicolumn{1}{c}{[m]} & \multicolumn{1}{c}{[sec]} 
       & \multicolumn{1}{c}{[\%]} & \multicolumn{1}{c}{[m]} & \multicolumn{1}{c}{[sec]} 
         \\ \midrule
& State-based (Song et al.~\cite{Song23Reaching})   & 
100.00 & 
0.516 & 
\underline{2.800} & 
100.00 & 
0.531 &
2.731 &
100.00 &
0.540 &
2.816
\\
& State-based (Song et al.~\cite{Song23Reaching}) +Perc    & 
100.00 & 
\underline{0.199} & 
2.810 & 
100.00 & 
\underline{0.196} &
\underline{2.729} &
100.00 &
\underline{0.221} &
2.817
\\
Ellipse
& State-based (ours)   & 
100.00 & 
0.296 & 
2.846 & 
100.00 & 
0.219 &
2.756 &
100.00 &
0.389 &
\underline{2.570}
\\
& Pixel-based (sym.) (ours)   & 
0.00 & 
\textbf{0.328} & 
- & 
0.00 & 
0.268 & 
- & 
0.00 & 
- &
- 
\\
& \textbf{Pixel-based (asym.) (ours)}   & 
\textbf{93.75} &
0.350 & 
\textbf{3.072} & 
\textbf{90.60} &
\textbf{0.154} & 
\textbf{2.902} & 
\textbf{100.00} &
\textbf{0.381} &
\textbf{3.318}
\\
\midrule 
& State-based (Song et al.~\cite{Song23Reaching})    & 
100.00 & 
0.446 & 
\underline{3.600} & 
100.00 & 
0.438 &
\underline{3.588} &
100.00 &
0.414 &
\underline{3.648}
\\
& State-based (Song et al.~\cite{Song23Reaching}) +Perc    & 
100.00 & 
\underline{0.190} & 
4.819 & 
100.00 & 
\underline{0.180} &
4.727 &
100.00 &
\underline{0.184} &
4.900
\\
Figure-8
& State-based (ours)& 
100.00 & 
0.200 & 
4.833 & 
100.00 & 
0.190 &
4.741 &
100.00 &
0.367 &
4.703
\\
& Pixel-based (sym.) (ours)  & 
0.00 & 
0.409 & 
- & 
0.00 & 
0.378 & 
- & 
0.00 & 
- &
- 
\\
& \textbf{Pixel-based (asym.) (ours)}   & 
\textbf{100.00} & 
\textbf{0.198} & 
\textbf{4.626} & 
\textbf{95.30} & 
\textbf{0.186} &
\textbf{4.644} &
\textbf{100.00} &
\textbf{0.238} &
\textbf{4.777}
\\
\midrule 
& State-based (Song et al.~\cite{Song23Reaching})    & 
100.00 & 
0.471 & 
\underline{3.479} & 
100.00 & 
0.451 &
\underline{3.497} &
100.00 &
0.455 &
\underline{3.570}
\\
& State-based (Song et al.~\cite{Song23Reaching}) +Perc    & 
100.00 & 
\underline{0.119} & 
5.065 & 
100.00 & 
\underline{0.121} &
5.018 &
100.00 &
\underline{0.163} &
5.090
\\
Glasses
& State-based (ours)   & 
100.00 & 
0.151 & 
5.102 & 
100.00 & 
0.163 &
4.985 &
100.00 &
0.191 &
5.157
\\
& Pixel-based (sym.) (ours)  & 
0.00 & 
0.402 & 
- & 
0.00 & 
0.396 & 
- & 
0.00 & 
- &
- 
\\
& \textbf{Pixel-based (asym.) (ours)}   & 
\textbf{100.00} & 
\textbf{0.240} & 
\textbf{5.267} & 
\textbf{89.10} & 
\textbf{0.209} &
\textbf{5.162} &
\textbf{100.00} &
\textbf{0.273} &
\textbf{5.586}
\\
\bottomrule
\end{tabularx}
\label{tab:main_results}
\end{table*}
After presenting the methodology in the previous section, we now aim to quantitatively evaluate the performance of our proposed system. The metrics used during this comparison will be (i) the success rate, quantifying how many runs successfully finish a three-lap race, (ii) the mean-gate-passing-error quantifying how far from the gate center the drone passes the gate, and (iii) the lap-time, measuring how agilely the control policy flies. Leveraging these metrics, we want to answer the following three research questions: (i) how does our pixel-based agent compare to a state-based agent on different track layouts, (ii) how sensitive is our agent to variations in the track layout and (iii) is our method able to robustly fly in the real world.

\subsection{Simulation \& HIL Experiments}
To obtain reproducible results, we conduct the first part of our analysis in simulation as well as using hardware-in-the-loop (HIL) experiments. During such HIL experiments the physical drone is controlled in the real world, however, instead of using the gate detector for segmentation, we use our simulated pixel-observations. This is possible since a motion-capture system gives us access to the pose of the drone in real time. This HIL approach enables differentiating between gate-detector performance and the performance of the RL agent.

To assess the effectiveness of our approach, we conduct a comparative analysis between our pixel-based policy and two baselines. The first baseline is a pixel-based policy which we have trained with a symmetric architecture, e.g. omitting the privileged information in the critic. The second baseline is a state-based policy where both actor and critic are trained exclusively on the full simulation state. Figure~\ref{fig:tracks} visualizes the training rewards of the three approaches together with the corresponding racetrack. From these plots, it is already obvious that the symmetric pixel-based agent is not able to fly the drone. The state-based agents learn much faster, thanks to the smaller and hightly task relevant observation space. This speed difference is especially pronounced when comparing training times, as the state-based agent is also eight times faster to train. However, given enough environment interactions, our asymmetric pixel-based agent performs similarly. For results on racetracks where no laps are flown (i.e. start to finish only), see appendix~\ref{sec:appendix_acyclic}. 

Table~\ref{tab:main_results} shows a detailed evaluation of the agents' performance on all three racetracks. We conduct the evaluation both in the BEM simulator, our augmented simulator (training environment), as well as in the real world with hardware-in-the-loop simulation. The simulation evaluation is conducted by simulating policy rollouts in each setting using 64 environments with slightly perturbed starting positions and 1000 time steps. For hardware in the loop, we run the policy multiple times and fly three laps during each run.

The results clearly show that the asymmetric critic is crucial for the method's success. Furthermore, our asymmetric pixel-based agents come very close to the performance of the state-based agents in all metrics. Most notably, it achieves a 100\% success rate during hardware-in-the-loop deployment of multiple experiments.

\begin{figure*}[ht]
    \centering
    \includegraphics{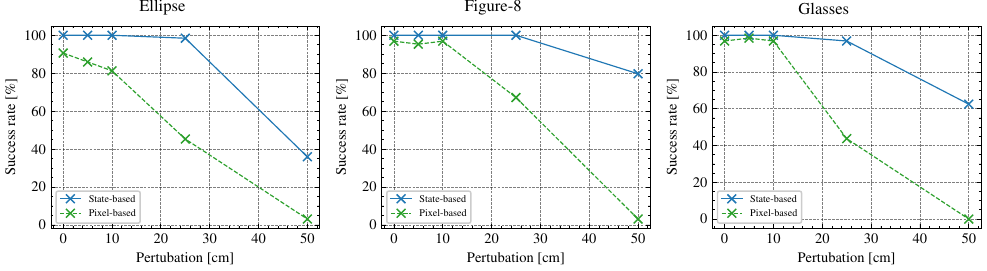}
\caption{The rate of successful trials in the augmented simulator ablated by perturbing the gate position in positive and negative $x, y, z$ direction by a uniform distribution. While the state-based policy benefits from direct access to gate position information, the pixel-based policy demonstrates robustness by effectively inferring gate position solely from image observations, even in scenarios deviating from the training environment.}
\label{fig:sensitivity}
\end{figure*}

In addition, we can notice that a state-based policy trained using the methodology of Song et al.~\cite{Song23Reaching} is superior in terms of lap time. However, these policies complete the tracks independently of their viewing direction, which gives them the advantage of not necessarily looking toward the gates the drone has to pass through. Moreover, the actions are not as regularized as our approach, resulting in more aggressive maneuvers, sharper turns, and noticeably higher mean-gate-passing-errors (MGE). The MGE can be interpreted as a measure of the safety margin when passing gates, as a higher deviation from the center of the gates makes crashes more likely but yields faster lap times. If we add the proposed perception-aware reward and action regularization to~\cite{Song23Reaching}, then the overall performance of~\cite{Song23Reaching} is comparable to our state-based policy (see +Perc entries). The remaining lap-time difference to the pixel-based policies is caused by ~\cite{Song23Reaching}+Perc being deployed with state-based information from a motion-capture system.

\subsubsection{Sensitivity to Gate Positions}
Next, we investigate the sensitivity of our pixel-based policy to slightly changed gate positions. By introducing gate position randomization, we aim to assess the generalizability of our policies across varying scenarios. This ablation provides valuable insights into how well our pixel-based approach adapts to changes in gate positions. In Figure~\ref{fig:sensitivity}, the results are visualized. At the beginning of each simulation experiment, the gates are randomly displaced by up to \unit[50]{cm} in each direction. Consistent with the methodology employed in previous simulation experiments, each metric is derived from the average of 64 environment rollouts, each spanning 1000 time steps. Notably, even without explicit gate position information, our pixel-based approach achieves successful flights, demonstrating resilience in the face of substantial gate-position perturbations. The state-based policy is superior because its input directly contains the ground-truth relative position of the next gate to be passed.

\subsection{Real-World Experiments}
In a final setup of experiments, we deploy our pixel-based policy together with the gate detector in the real world on the Figure 8 track.
We use a modification of the \emph{Agilicious} platform \cite{Foehn22Agi} for the real-world deployment. 
On this modified platform, the onboard computer is replaced with an RF receiver which receives the control commands from the ground station. The RF receiver is connected to the flight controller\footnote{\url{https://www.betaflight.com}}, which then executes the transmitted collective thrust and body rate commands.
Additionally, the quadrotor is equipped with a low-latency video transmission system which sends the live video stream to the base computer. 
This configuration is similar to the one used by professional drone racing pilots and we measure at video stream latency of \unit[33]{ms}. 
The gate detector is run directly on the images sent to the ground station.

Table~\ref{tab:real_results} summarizes the real-world flights. For a more dynamic impression, the reader is advised to watch the supplementary video showcasing these experiments. We achieve direct zero-shot simulation-to-reality transfer of our pixel-based control policies. The success rate across 6 runs and a total of 20 flown laps is 100\%. Compared to the state-based policy, the asymmetric pixel-based policy exhibits a slightly larger gate-passing error but flies a similar lap-time.

This result means that we have demonstrated agile flight directly from pixels without explicit state estimation, as our asymmetric pixel-based agent robustly navigates the track at speeds up to \unit[40]{km/h} and accelerations up to \unit[2]{g}.

\begin{table}
\centering
\footnotesize
\caption{\textnormal{Results in the real world flying without state estimation for 6 trials with 3 laps each on the Figure-8 racetrack. We compare the success rate~(SR), mean-gate-passing-error~(MGE), and the lap-time~(LT) in the real-world using state-based and pixel-based observations.}}
\begin{tabularx}{0.45\textwidth}{l c *2{C}}
\toprule
        & \multicolumn{1}{c}{SR} & \multicolumn{1}{c}{MGE} & \multicolumn{1}{c}{LT}  \\
 Observation & \multicolumn{1}{c}{[\%]} & \multicolumn{1}{c}{[m]} & \multicolumn{1}{c}{[sec]}
         \\ \midrule
State-based    & 
100.00 & 
0.367 &
4.703
\\
\textbf{Pixel-based (asym.)}   & 
100.00 &
0.491 &
4.683 
\\
\bottomrule
\end{tabularx}
\label{tab:real_results}
\end{table}

\pagebreak

\section{Discussion and Conclusion}\label{sec:discussion}
In this paper, we presented, to our knowledge, the first vision-based, agile quadrotor system that learns to directly map pixels to low-level control commands without explicit state estimation or access to IMU.
This combination of observation and action space has been a long-standing milestone for visuomotor robotic intelligence.
Through extensive experiments in simulation and real-world drone racing at speeds up to 40\,km/h with accelerations up to 2\,g, our methodology showcases the direct transfer of policies from simulation to reality with the appropriate abstractions of the visual input feed, akin to professional human drone pilots.

As shown in the experiments, the asymmetric actor-critic architecture plays a vital role in the approach's success. By leveraging privileged information about the drone's state in the environment, the critic network enhances its precision in assessing the actions taken by the actor, particularly in the context of pixel observations. This framework enables performance similar to a state-based policy, achieving comparable but marginally lower success rates in our evaluations. The inner gate edge abstraction enables training policies with up to 400 million environment interactions within a day. This makes it possible to train a pixel-based agent directly with RL, contributing to the robustness of the agent. 

At present, a primary limitation of our approach is the limited memory of the neural controller, only considering three past actions. When the drone is oriented so that no gate is visible for multiple frames, the success rate drops drastically as our architecture's relatively short action history hinders recovery. This challenge can be effectively addressed by incorporating a recurrent architecture, allowing for prolonged memory retention. Furthermore, enhancing sample efficiency could be achieved by employing specialized algorithms tailored for learning from pixels.

While this work exclusively demonstrates autonomous vision-based flight without state-estimation for the task of drone racing, we believe that it has broader implications and will serve as a foundation for more application-oriented extensions. In tasks like autonomous indoor navigation or ship inspection, salient landmarks like doorways or manholes could be used instead of gates as their positions are known and their appearance is relatively consistent. Here, employing a separate the vision-encoder and RL-controller is beneficial as the landmark detector can be trained independently of the navigation policy. The latter can be trained in a simulation with access to a floorplan or blueprint of the deployment environment, potentially enabling diverse inspection tasks.

Another potential application of our method is to overcome situations where IMU information is not reliable, such as a straight-line flight at a constant speed. This flight profile, prevalent during powerline inspection, renders the IMU bias drift in a state-estimation pipeline unobservable. Our method on the other hand could rely on detecting powerline pylons and always fly towards the next visible pylon, effectively traversing along the powerline. 

We realize that future research dedicated to obstacle avoidance, dynamic objects and unforseen changes to the environment is required for many real-world applications. Nevertheless, we believe that our work represents a major step forward in the development and understanding of fully autonomous robots that rely solely on visual data.

\section*{Acknowledgments}
This work was supported by the European Union’s Horizon Europe Research and Innovation Programme under grant agreement No. 101120732 (AUTOASSESS) and the European Research Council (ERC) under grant agreement No. 864042 (AGILEFLIGHT). The authors thank Chunwei Xing for his assistance in the development of the Swin-transformer-based gate detector.

{
\small
\bibliographystyle{unsrtnat}
\bibliography{main}
}
\clearpage

\begin{figure}[h!]  
    \centering
    \includegraphics[width=2.04\linewidth]{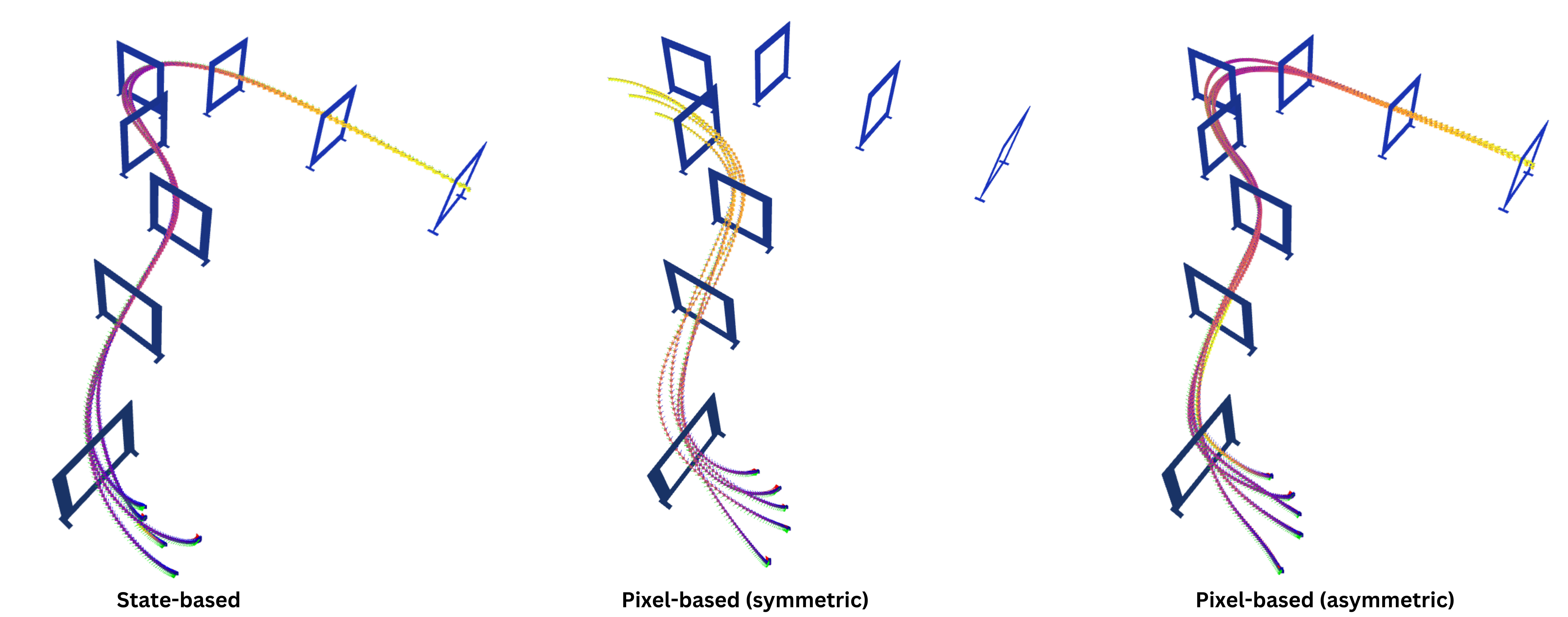}
    \parbox{2.04\linewidth}{\caption{Comparison of multiple rollouts with different initial conditions on an acyclic racetrack. From left to right: rollouts with a state-based policy, a pixel-based symmetric actor-critic policy, and the proposed pixel-based asymmetric architecture. The symmetric pixel-based policy is not able to learn this racetrack successfully.} \label{fig:acyclic}}
\end{figure}

\section{Appendix}\label{sec:appendix}
\subsection{Acyclic racetrack}\label{sec:appendix_acyclic}

In a drone-racing scenario, it is a common assumption that the race lasts multiple laps; thus, the task becomes cyclic. However, outside of drone racing, a drone might be confronted with an acyclic task consisting of flying from one point to another. To underline that our method also allows for this type of mission, we also evaluate it on an acyclic racetrack, where the episode terminates after the last gate. Besides this change, the observations, rewards, and actions remain the same. A racecourse without cycles demonstrates that this method is not limited to drone racing and can, therefore, generalize beyond other maps that are used for vision-based navigation.
Notice that the symmetric actor-critic architecture failed to complete this track entirely. In contrast, the proposed asymmetric architecture successfully navigated the track in multiple runs with varying starting positions, similar to the state-based policy, see Figure~\ref{fig:acyclic}.

\begin{figure}[!h]
    \centering
    \includegraphics[scale=0.95]{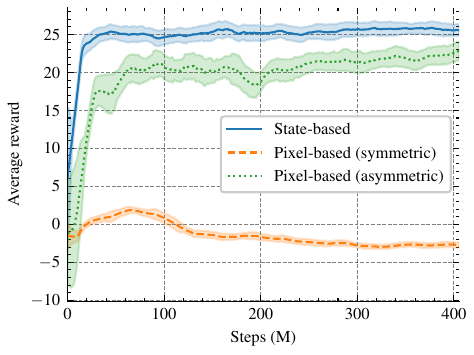}
    \caption{Reward progress over the number of simulation time-steps for the acyclic track. The symmetric actor-critic is not able to complete the whole track.}
    \label{fig:acyclic_rew}
\end{figure}

The corresponding average reward while training the acyclic track is visualized in Figure~\ref{fig:acyclic_rew}. Since the episode terminates after the last gate with a terminal reward of $r_t^\text{term} = 10$, the total reward is lower than the previously discussed cyclic racetracks reported in Figure~\ref{fig:tracks}, where episodes could go up to 30 seconds (assuming no collision before) with a simulation step frequency of 50 Hz.

\subsection{Hyperparameters}

We train our state- and vision-based policies using Proximal Policy Optimization (PPO), simulating 100 environments in parallel to collect rollouts. The hyperparameters listed in Table~\ref{tab:hyperparameters} are used to train all the policies. The only exception is the acyclic track, which requires a lower discount factor $\gamma = 0.98$ for robust learning, since the episodes are significantly shorter than tracks where the drone may complete multiple laps in up to 30 seconds.

\newpage
\vspace*{8cm}

\begin{table}[!h]
\centering
\caption{\textnormal{PPO hyperparameters to train state- and vision-based policies.}}
\scriptsize
\setlength{\tabcolsep}{12pt}
\begin{tabularx}{0.4\textwidth}{ll}
\toprule
   Parameter & Value \\ \midrule
learning rate & 3e-4 linear decay to 1e-5 \\
discount factor & 0.995 \\
GAE-$\lambda$ & 0.95 \\
learning epochs & 10 \\ 
clip range & 0.2 \\
entropy coefficient & 0.001 \\ 
batch size & 25000 \\
policy network MLP & [512, 512] \\
value network MLP & [512, 512] \\
CNN-encoder latent dimension & 256\\
\bottomrule
\end{tabularx}
\label{tab:hyperparameters}
\end{table}

\end{document}